\documentclass[11pt,a4paper]{article}
\usepackage{times,latexsym}
\usepackage{url}
\usepackage[T1]{fontenc}

    \usepackage[acceptedWithA]{tacl2018v2}
\usepackage[]{tacl2018v2}

\usepackage{amsmath}
\usepackage{multirow}
\usepackage{graphicx}  
\usepackage{pinyin}
\usepackage{CJK} 
\usepackage{booktabs}
\usepackage[normalem]{ulem}
\usepackage{color}
\usepackage{amsfonts}
\usepackage[english]{babel}

\usepackage{xspace,mfirstuc,tabulary}

\newif\iftaclinstructions
\taclinstructionsfalse 
\iftaclinstructions
\renewcommand{\confidential}{}
\renewcommand{\anonsubtext}{(No author info supplied here, for consistency with
TACL-submission anonymization requirements)}
\newcommand{\instr}
\fi

\iftaclpubformat 

\else

\fi

\title{Synchronous Bidirectional Neural Machine Translation}

\author{Long Zhou$^{1,2}$, Jiajun Zhang$^{1,2}$\Thanks{Corresponding author.}, Chengqing Zong$^{1,2,3}$\\
	$^1$National Laboratory of Pattern Recognition, CASIA, Beijing, China \\
	$^2$University of Chinese Academy of Sciences, Beijing, China \\
	$^3$CAS Center for Excellence in Brain Science and Intelligence Technology, Shanghai, China \\
	{\sf \{long.zhou, jjzhang, cqzong\}@nlpr.ia.ac.cn }\\
}

\date{}

\begin{document}
\begin{CJK*}{UTF8}{gbsn}

\maketitle

\begin{abstract}
Existing approaches to neural machine translation (NMT) generate the target language sequence token by token from left to right. However, this kind of unidirectional decoding framework cannot make full use of the target-side future contexts which can be produced in a right-to-left decoding direction, and thus suffers from the issue of unbalanced outputs.
In this paper, we introduce a synchronous bidirectional neural machine translation (SB-NMT) that predicts its outputs using left-to-right and right-to-left decoding simultaneously and interactively, in order to leverage both of the history and future information at the same time.
Specifically, we first propose a new algorithm that enables synchronous bidirectional decoding in a single model. Then, we present an interactive decoding model in which left-to-right (right-to-left) generation does not only depend on its previously generated outputs, but also relies on future contexts predicted by right-to-left (left-to-right) decoding.
We extensively evaluate the proposed SB-NMT model on large-scale NIST Chinese-English, WMT14 English-German, and WMT18 Russian-English translation tasks.
Experimental results demonstrate that our model achieves significant improvements over the strong Transformer model by 3.92, 1.49 and 1.04 BLEU points respectively, and obtains the state-of-the-art performance on Chinese-English and English-German translation tasks.{\footnote[1]{The source code is available at \url{https://github.com/wszlong/sb-nmt}.}}
\end{abstract}

\section{Introduction}

Neural machine translation has significantly improved the quality of machine translation in recent years~\cite{Sutskever:2014,Bahdanau:2015,Zhang:2015,Wu:2016, gehring2017convolutional,vaswani2017attention}.
Recent approaches to sequence to sequence learning typically leverage recurrence ~\cite{Sutskever:2014}, convolution~\cite{gehring2017convolutional}, or attention ~\cite{vaswani2017attention} as basic building blocks.

\begin{table}
\centering
\begin{tabular}{|l||c|c|}
  \hline
   Model              &    The first 4 tokens &  The last 4 tokens               \\
  \hline
  \hline
   L2R               &      \textbf{40.21}\%       &   35.10\%       \\
   R2L                &      35.67\%       &   \textbf{39.47\%}      \\
   \hline
\end{tabular}
\caption{Translation accuracy of the first 4 tokens and last 4 tokens in NIST Chinese-English translation tasks.
  L2R denotes left-to-right decoding and R2L means right-to-left decoding for conventional NMT.
  } \label{trans-acc}
\end{table}

Typically, NMT adopts the encoder-decoder architecture and generates the target translation from left to right. Despite their remarkable success, NMT models suffer from several weaknesses~\cite{koehn2017six}.
One of the most prominent issues is the problem of unbalanced outputs in which the translation prefixes are better predicted than the suffixes~\cite{liu2016agreementa}.
We analyze translation accuracy of the first and last 4 tokens for left-to-right (L2R) and right-to-left (R2L) directions respectively.
As shown in Table~\ref{trans-acc}, the statistical results show that L2R performs better in the first 4 tokens, whereas R2L translates better in term of the last 4 tokens.
This problem is mainly caused by the left-to-right unidirectional decoding, which conditions each output word on previously generated outputs only, but leaving the future information from target-side contexts unexploited during translation.
The future context is commonly used in reading and writing in human cognitive process~\cite{NIPS2017_6775}, and it is crucial to avoid under-translation~\cite{Tu:2016,D16-1096}.

To alleviate the problems, existing studies usually used independent bidirectional decoders for NMT~\cite{liu2016agreementa,W16-2323}.
Most of them trained two NMT models with left-to-right and right-to-left directions respectively. Then, they translated and re-ranked candidate translations using two decoding scores together.
More recently, \newcite{zhang2018asynchronous} presented an asynchronous bidirectional decoding algorithm for NMT, which extended the conventional encoder-decoder framework by utilizing a backward decoder.
However, these methods are more complicated than the conventional NMT framework beacuse they require two NMT models or decoders.
Furthermore, the L2R and R2L decoders are independent from each other~\cite{liu2016agreementa}, or only the forward decoder can utilize information from the backward decoder~\cite{zhang2018asynchronous}.
It is therefore a promising direction to design a synchronous bidirectional decoding algorithm in which L2R and R2L generations can interact with each other.

\begin{figure}
    \centering
    \includegraphics[width=7.5cm]{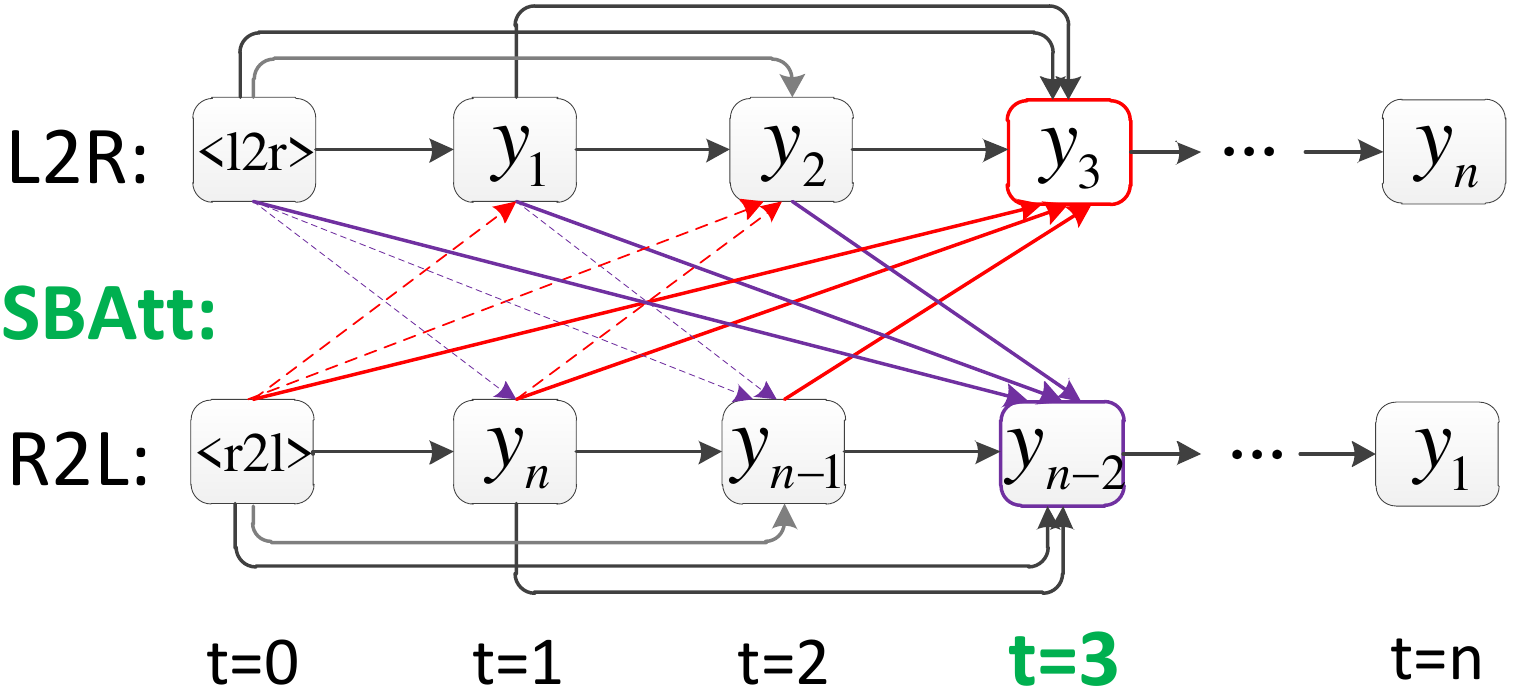}
    \caption{Illustration of the decoder in the synchronous bidirectional NMT model. L2R denotes left-to-right decoding guided by the start token $\langle l2r \rangle$ and R2L means right-to-left decoding indicated by the start token $\langle r2l \rangle$.
    SBAtt is our proposed synchronous bidirectional attention (see \S~\ref{SBA-sec}).
    For instance, the generation of $y_3$ does not only rely on $y_1$ and $y_2$, but also depends on $y_n$ and $y_{n-1}$ of R2L.
    }\label{Bi-generation}
\end{figure}

Accordingly, we propose in this paper a novel framework (SB-NMT) that utilizes a single decoder to bidirectionally generate target sentences simultaneously and interactively.
As shown in Figure~\ref{Bi-generation}, two special labels ($\langle l2r \rangle$ and $\langle r2l \rangle$) at the beginning of the target sentence guide translating from left to right or right to left, and the decoder in each direction can utilize the previously generated symbols of bidirectional decoding when generating the next token.
Taking L2R decoding as an example, at each moment, the generation of the target word (e.g., $y_3$) does not only rely on previously generated outputs ($y_1$ and $y_2$) of L2R decoding, but also depends on previously predicted tokens ($y_n$ and $y_{n-1}$) of R2L decoding. 
Compared to the previous related NMT models, our method has the following advantages:
1) We use a single model (one encoder and one decoder) to achieve the decoding with left-to-right and right-to-left generation, which can be processed in parallel.
2) Via the synchronous bidirectional attention model (SBAtt, \S\ref{SBA-sec}), our proposed model is an end-to-end joint framework and can optimize bidirectional decoding simultaneously.
3) Compared to two-phase decoding scheme in previous work, our decoder is faster and more compact using one beam-search algorithm.

Specifically, we make the following contributions in this paper:
\begin{itemize}
\item We propose a synchronous bidirectional NMT model that adopts one decoder to generate outputs with left-to-right and right-to-left directions simultaneously and interactively.
To the best of our knowledge, this is the first work to investigate the effectiveness of a single NMT model with synchronous bidirectional decoding.
\item Extensive experiments on NIST Chinese-English, WMT14 English-German and WMT18 Russian-English translation tasks demonstrate that our SB-NMT model obtains significant improvements over the strong Transformer model by 3.92, 1.49 and 1.04 BLEU points respectively.
In particular, our approach separately establishes the state-of-the-art BLEU score of 51.11 and 29.21 on Chinese-English and English-German translation tasks.
\end{itemize}

\section{Background}

In this paper, we build our model based on the powerful Transformer~\cite{vaswani2017attention} with an encoder-decoder framework, where the encoder network first transforms an input sequence of symbols $x=(x_1,x_2,...,x_n)$ to a sequence of continues representations $z=(z_1,z_2,...,z_n)$, from which the decoder generates an output sequence $y=(y_1,y_2,...,y_m)$ one element at a time.
Particularly, relying entirely on the multi-head attention mechanism, the Transformer with beam search algorithm achieves the state-of-the-art results for machine translation.

\begin{figure}
    \centering
    \includegraphics[height=5cm]{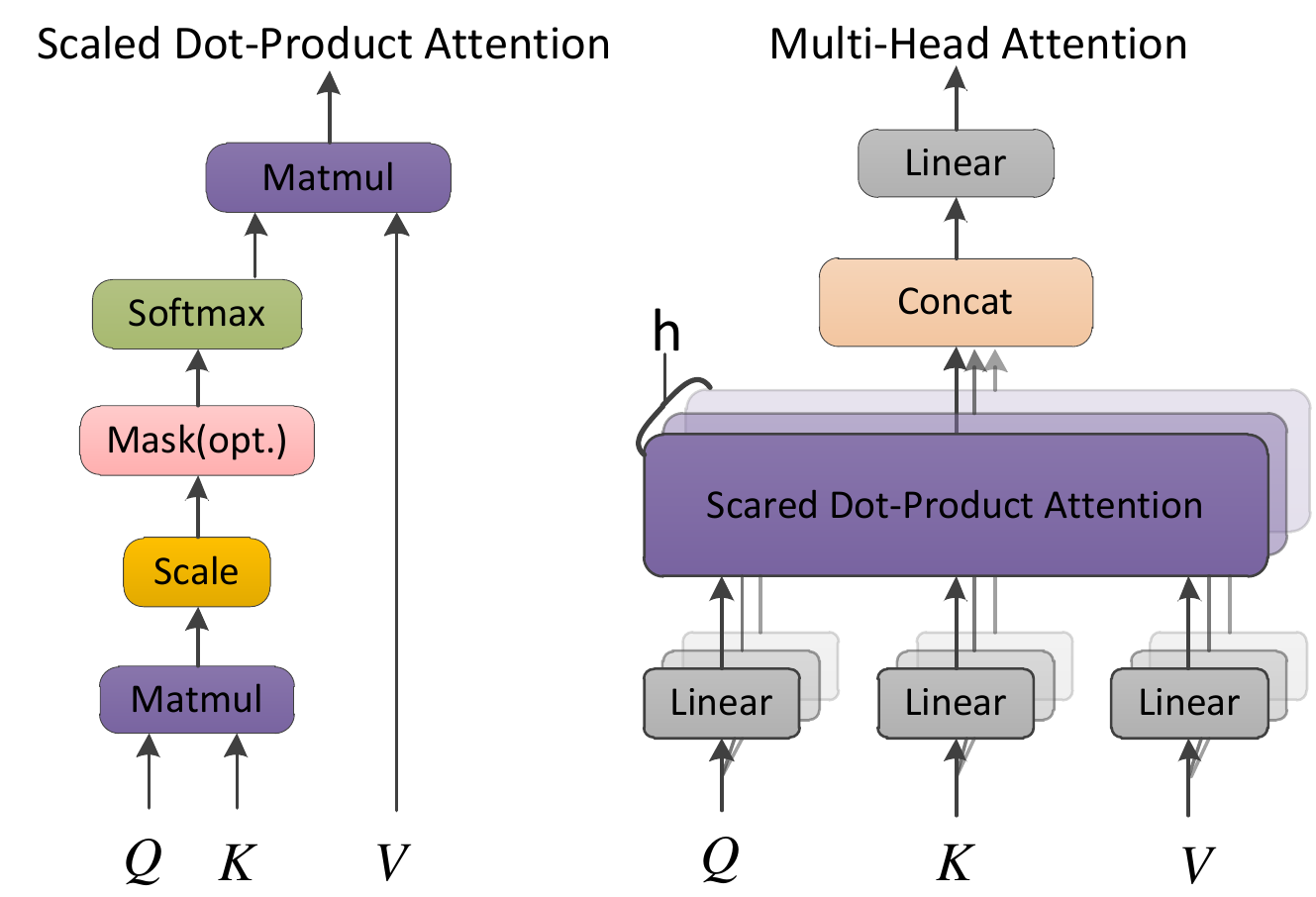}
    \caption{(left) Scaled Dot-Product Attention. (right) Multi-Head Attention.
    }\label{Dot-Procdut}
\end{figure}

~\textbf{Multi-Head Attention} allows the model to jointly attend to information from different representation subspaces at different positions.
It operates on queries $Q$, keys $K$, and values $V$.
For multi-head intra-attention of encoder or decoder, all of $Q, K, V$ are the output hidden state matrices of the previous layer. For multi-head inter-attention of the decoder, Q are the hidden states of the previous decoder layer, and $K$-$V$ pairs come from the output $(z_1, z_2,..., z_n)$ of the encoder.

Formally, multi-head attention first obtains $h$ different representations of ($Q_i,K_i,V_i$). Specifically, for each attention head $i$, we project the hidden state matrix into distinct query, key and value representations $Q_i$=$QW_i^Q$, $K_i$=$KW_i^K$,  $V_i$=$VW_i^V$ respectively.
Then we perform {\bf scaled dot-product attention} for each representation, concatenate the results, and project the concatenation
with a feed-forward layer.  

\begin{equation}
	\begin{aligned} \label{MHAtt}
		\mbox{MultiHead}(Q,K,V) = \mbox{Concat}_i(head_i)W^O  \\
		head_i = \mbox{Attention}(QW_i^Q, KW_i^K, VW_i^V)
	\end{aligned}
\end{equation}
where $W_i^Q$, $W_i^K$, $W_i^V$ and $W^O$ are parameter projection matrices .

~\textbf{Scaled Dot-Product Attention} can be described as mapping a query and a set of key-value pairs to an output. 
Specifically, we can then multiply query $Q_i$ by key $K_i$ to obtain an attention weight matrix, which is then multiplied by value $V_i$ for each token to obtain the self-attention token representation.
As shown in Figure~\ref{Dot-Procdut}, scaled dot-product attention operates on a query $Q$, a key $K$, and a value $V$ as:
\begin{equation}
    \mbox{Attention}(Q,K,V) = \mbox{Softmax}(\frac{QK^T}{\sqrt{d_k}})V   \label{attention}
\end{equation}
where $d_k$ is the dimension of the key. For the sake of brevity, we refer the reader to \newcite{vaswani2017attention} for more details.

\begin{figure}
	\centering
	\includegraphics[height=6cm]{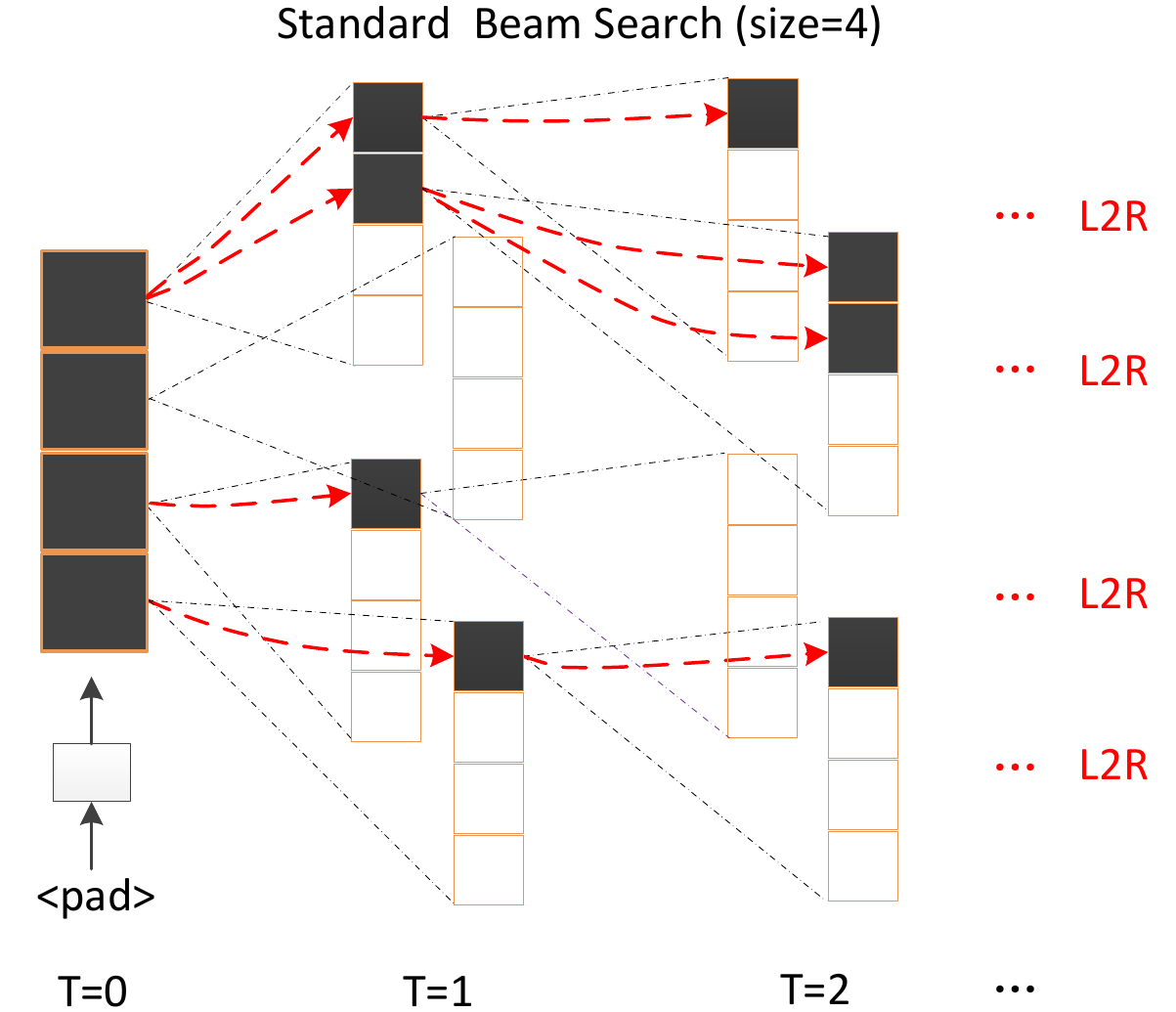}
	\caption{Illustration of the standard beam search algorithm with beam size 4. The black blocks denote the ongoing expansion of the hypotheses.
	}\label{decoding-std}
\end{figure}

~\textbf{Standard Beam Search}
Given the trained model and input sentence $x$, we usually employ beam search or greedy search (beam size = 1) to find the best translation $\widehat{y}={\rm{argmax}}_yP(y|x)$.
Beam size N is used to control the search space by extending only the top-N hypotheses in the current stack.
As shown in Figure~\ref{decoding-std}, the blocks represent the four best token expansions of the previous states, and these token expansions are sorted top-to-bottom from most-probable to least-probable.
We define a complete hypothesis as a hypothesis which outputs EOS, where EOS is a special target token indicating the end of sentence.
With the above settings, the translation $y$ is generated token by token from left to right.

\section{Our Approach}

In this section, we will introduce the approach of synchronous bidirectional NMT. Our goal is to design a synchronous bidirectional beam search algorithm  ($\S\ref{SBBS-sec}$) which generates tokens with both L2R and R2L decoding simultaneously and interactively using a single model.
The central module is the synchronous bidirectional attention (SBAtt, see \S\ref{SBA-sec}). By using SBAtt, the two decoding directions in one beam-search process can help and interact with each other, and can make full use of the target-side history and future information during translation.
Then, we apply our proposed SBAtt to replace the multi-head intra-attention in the decoder part of Transformer model (\S\ref{Integrating-sec}),
and the model is trained end-to-end by maximum likelihood using stochastic gradient descent ($\S\ref{Training-sec}$).

\begin{figure}
	\centering
	\includegraphics[height=6.5cm]{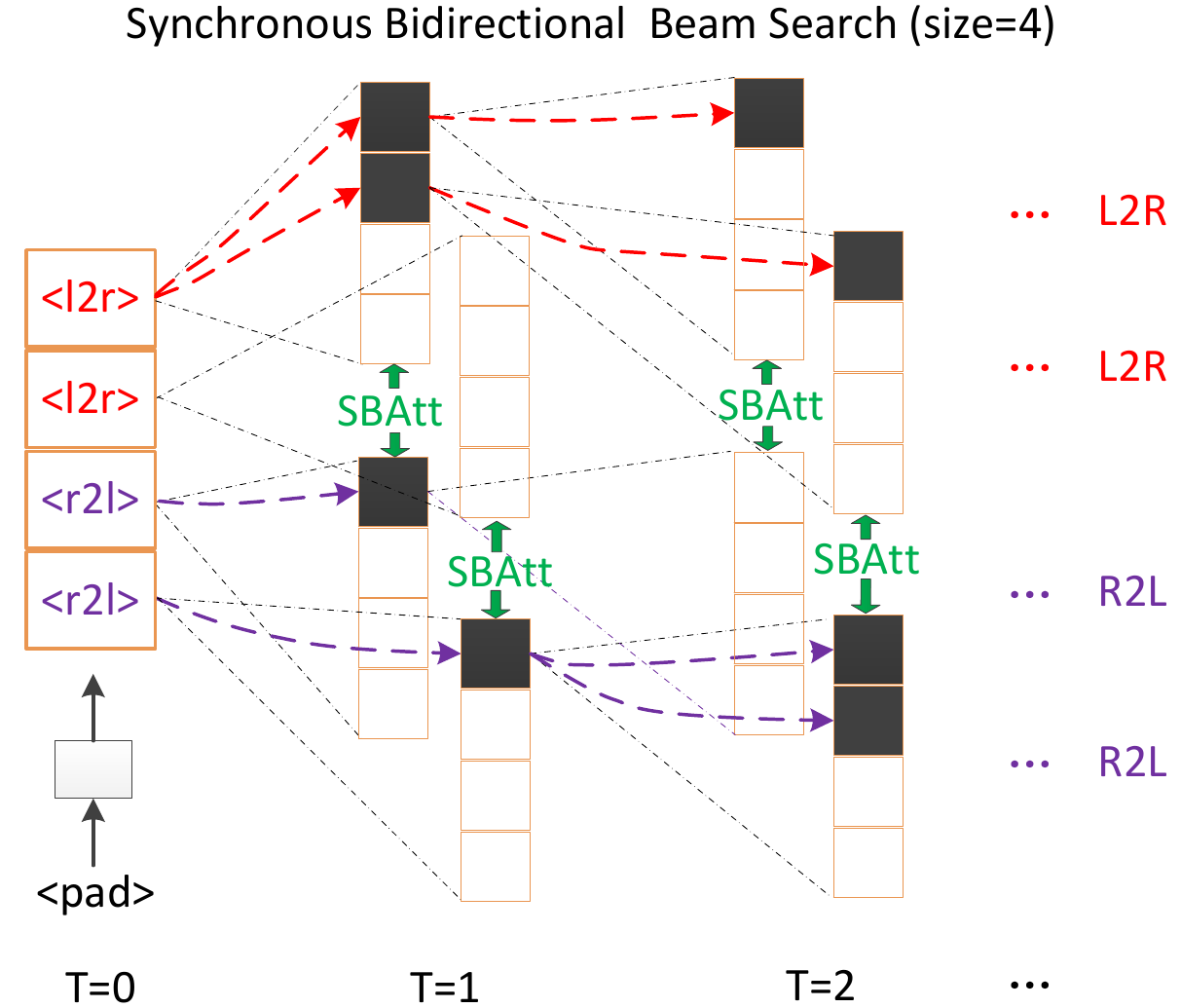}
	\caption{The synchronous bidirectional decoding of our model. $\langle${l2r}$\rangle$ and $\langle${r2l}$\rangle$ are two special labels, which indicate the target-side translation direction in L2R and R2L modes, respectively.
	Our model can decode with both L2R and R2L directions in one beam search by using SBAtt, simultaneously and interactively.
	SBAtt means the synchronous bidirectional attention (\S\ref{SBA-sec}) performed between items of L2R and R2L decoding.
	}\label{decoding}
\end{figure}

\subsection{Synchronous Bidirectional Beam Search} \label{SBBS-sec}
Figure~\ref{decoding} illustrates the synchronous bidirectional beam-search process with beam size 4.
With two special start tokens which are optimized during the training process, we let half of the beam to keep decoding from left to right guided by the label $\langle${l2r}$\rangle$, and allow the other half beam to decode from right to left indicated by the label $\langle${r2l}$\rangle$.
More importantly, via the proposed \textbf{SBAtt} (\S\ref{SBA-sec}) model, L2R (R2L) generation does not only depend on its previously generated outputs, but also relies on future contexts predicted by R2L (L2R) decoding.

Note that 
(1) at each time step, we choose best items of the half beam from L2R decoding and best items of the half beam from R2L decoding to continue expanding simultaneously;
(2) L2R and R2L beams should be thought of as parallel, with \textbf{SBAtt} computed between items of 1-best L2R and R2L, items of 2-best L2R and R2L, and so on{\footnote[2]{We also did experiments that all of L2R hypotheses attend to the 1-best R2L hypothesis, and all the R2L hypotheses attend to the 1-bset L2R hypothesis. The results of the two schemes are similar. For the sake of simplicity, we employed the previous scheme.}};
(3) the black blocks denote the ongoing expansion of the hypotheses and decoding terminates when the end-of-sentence flag EOS is predicted;
(4) in our decoding algorithm, the complete hypotheses will not participate in subsequent SBAtt, and the L2R hypothesis attended by R2L decoding may change at different time steps, while the ongoing partial hypotheses in both directions of SBAtt always share the same length;
(5) finally, we output the translation result with highest probability from all complete hypotheses.
Intuitively, our model is able to choose from L2R output or R2L output as final hypothesis according to their model probabilities, and if a R2L hypothesis wins, we reverse the tokens before presenting it.

\subsection{Synchronous Bidirectional Attention} \label{SBA-sec}

Instead of multi-head intra-attention which prevents future information flow in the decoder to preserve the auto-regressive property, we propose a synchronous bidirectional attention (SBAtt) mechanism.
With the two key modules of synchronous bidirectional dot-product attention (\S\ref{SBDPA-sec}) and synchronous bidirectional multi-head attention (\S\ref{SBMHA-sec}), SBAtt is capable of capturing and combining the information generated by L2R and R2L decoding.
 
\begin{figure}
    \centering
    \includegraphics[height=7cm]{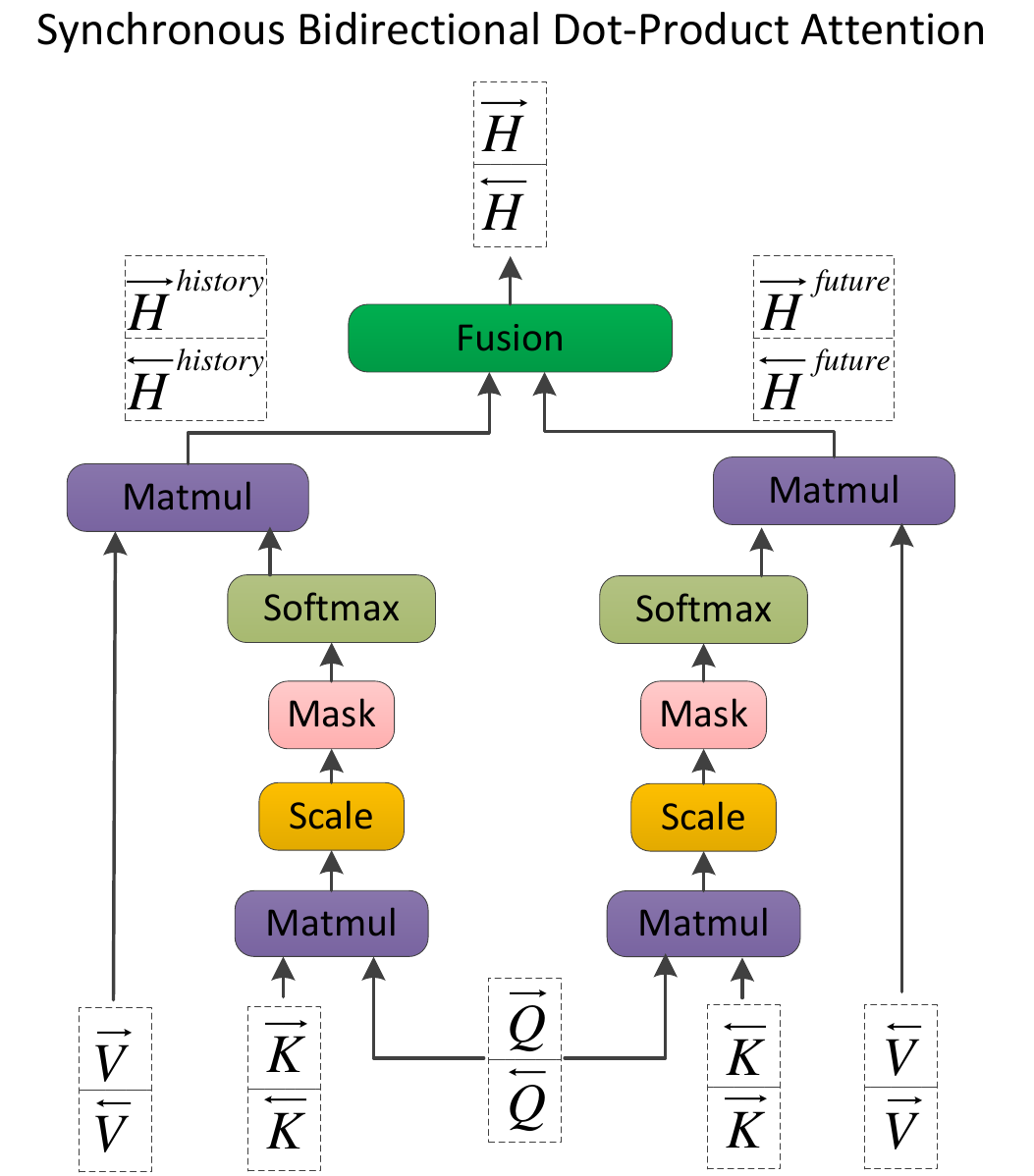}
    \caption{Synchronous bidirectional attention model based on scaled dot-product attention.
    It operates on forward (L2R) and backward (R2L) queries Q, keys K, values V.
    }\label{Bi-att}
\end{figure}

\subsubsection{Synchronous Bidirectional Dot-Product Attention} \label{SBDPA-sec}

Figure~\ref{Bi-att} shows our particular attention ``Synchronous Bidirectional Dot-Product Attention (SBDPA)''.
The input consists of queries ([$\overrightarrow{Q}$;$\overleftarrow{Q}$]), keys ([$\overrightarrow{K}$;$\overleftarrow{K}$]) and values ([$\overrightarrow{V}$;$\overleftarrow{V}$])
which are all concatenated by forward (L2R) states and backward (R2L) states.
The new forward state $\overrightarrow{H}$ and backward state $\overleftarrow{H}$ can be obtained by synchronous bidirectional dot-product attention. For the new forward state $\overrightarrow{H}$, it can be calculated as:
\begin{equation}\label{F-SBA}
\begin{aligned}
  \overrightarrow{H}^{history} = \mbox{Attention}(\overrightarrow{Q}, \overrightarrow{K}, \overrightarrow{V}) \\
  \overrightarrow{H}^{future} = \mbox{Attention}(\overrightarrow{Q}, \overleftarrow{K}, \overleftarrow{V}) \\
  \overrightarrow{H} = \mbox{Fusion}(\overrightarrow{H}^{history}, \overrightarrow{H}^{future})
\end{aligned}
\end{equation}
where $\overrightarrow{H}^{history}$ is obtained by using conventional scaled dot-product attention as introduced in Equation~\ref{attention}, and its purpose is to take advantage of previously generated tokens, namely {\bf history information}.
We calculate $\overrightarrow{H}^{future}$ using forward query ($\overrightarrow{Q}$) and backward key-value pairs ($\overleftarrow{K}$, $\overleftarrow{V}$), which attempts at making use of {\bf future information} from R2L decoding as effectively as possible in order to help predict the current token in L2R decoding.
The role of $\mbox{Fusion}(\cdot)$ (green block in Figure~\ref{Bi-att}) is to combine $\overrightarrow{H}^{history}$ and $\overrightarrow{H}^{future}$ by using linear interpolation, nonlinear interpolation or gate mechanism.

~\textbf{Linear Interpolation} $\overrightarrow{H}^{history}$ and $\overrightarrow{H}^{future}$ have different importance to prediction of current word. Linear interpolation of $\overrightarrow{H}^{history}$ and $\overrightarrow{H}^{future}$ produces an overall hidden state:
\begin{equation} \label{Linear}
  \overrightarrow{H} = \overrightarrow{H}^{history} + \lambda * \overrightarrow{H}^{future}
\end{equation}
where $\lambda$ is a hyper-parameter decided by the performance on development set.{\footnote[3]{Note that we can also set $\lambda$ to be a vector and learn $\lambda$ during training with standard back-propagation, and we remain it as future exploration.}}

~\textbf{Nonlinear Interpolation} $\overrightarrow{H}$ is equal to $\overrightarrow{H}^{history}$ in the conventional attention mechanism, and $\overrightarrow{H}^{future}$ means the attention information between current hidden state and generated hidden states of the other decoding. In order to distinguish two different information sources, we present a nonlinear interpolation by adding an activation function to the backward hidden states:
\begin{equation} \label{Nonlinear}
  \overrightarrow{H} = \overrightarrow{H}^{history} + \lambda * AF(\overrightarrow{H}^{future})
\end{equation}
where $AF$ denotes activation function, such as $tanh$ or $relu$.

~\textbf{Gate Mechanism} We also propose a gate mechanism to dynamically control the amount of information flow from the forward and backward contexts. Specially, we apply a feed-forward gating layer upon $\overrightarrow{H}^{history}$ as well as $\overrightarrow{H}^{future}$ to enrich the non-linear expressiveness of our model:
\begin{equation} \label{Gate}
\begin{aligned}
  r_t, z_t &= \sigma(W^g[\overrightarrow{H}^{history};\overrightarrow{H}^{future}])  \\
  \overrightarrow{H} &= r_t \odot \overrightarrow{H}^{history} + z_t \odot \overrightarrow{H}^{future}
\end{aligned}
\end{equation}
where $\odot$ denotes element-wise multiplication.
Via this gating layer, it is able to control how much past information can be preserved from previous context and how much reversed information can be captured from backward hidden states.

Similar to the calculation of forward hidden states $\overrightarrow{H}_{i}$, the backward hidden states $\overleftarrow{H}_{i}$ can be computed as follows.
\begin{equation}\label{B-SBA}
\begin{aligned}
  \overleftarrow{H}^{history} = \mbox{Attention}(\overleftarrow{Q}, \overleftarrow{K}, \overleftarrow{V}) \\
  \overleftarrow{H}^{future} = \mbox{Attention}(\overleftarrow{Q}, \overrightarrow{K}, \overrightarrow{V}) \\
  \overleftarrow{H} = \mbox{Fusion}(\overleftarrow{H}^{history}, \overleftarrow{H}^{future})
\end{aligned}
\end{equation}
where $\mbox{Fusion} (\cdot)$ is the same as introduced in Equation~\ref{Linear}-\ref{Gate}. Note that $\overrightarrow{H}$ and $\overleftarrow{H}$ can be calculated in parallel.
We refer to the whole procedure formulated in Equation~\ref{F-SBA} and Equation~\ref{B-SBA} as SBDPA($\cdot$).
\begin{equation}\label{SBAtt}
\begin{aligned}
  [][ \overrightarrow{H};\overleftarrow{H}] = \mbox{SBDPA}([\overleftarrow{Q};\overrightarrow{Q}],[\overleftarrow{K};\overrightarrow{K}], [\overleftarrow{V};\overrightarrow{V}])
\end{aligned}
\end{equation}

\subsubsection{Synchronous Bidirectional Multi-Head Attention} \label{SBMHA-sec}
Multi-head attention consists of $h$ attention heads, each of which learns a distinct attention function to attend to all of the tokens in the sequence, where mask is used for preventing leftward information flow in decoder.
Compared to the multi-head attention, our inputs are the concatenation of forward and backward hidden states.
We extend standard multi-headed attention by letting each head attend to both forward and backward hidden states, combined via SBDPA($\cdot$).

\begin{equation}
\begin{aligned}
    \mbox{MultiHead}([\overleftarrow{Q};\overrightarrow{Q}],[\overleftarrow{K};\overrightarrow{K}],[\overleftarrow{V};\overrightarrow{V}]) \\
     = \mbox{Concat}([\overrightarrow{H}_{1};\overleftarrow{H}_{1}],...,[\overrightarrow{H}_{h};\overleftarrow{H}_{h}])W^O  \\
\end{aligned}
\end{equation}
and $[\overrightarrow{H}_{i};\overleftarrow{H}_{i}]$ can be computed as follows, which is the biggest difference from conventional multi-head attention.
\begin{equation}
\begin{aligned}
    [][\overrightarrow{H}_{i};\overleftarrow{H}_{i}] = {\mbox{SBDPA}}([\overleftarrow{Q};\overrightarrow{Q}]{W_i^Q}, \\
        [\overleftarrow{K};\overrightarrow{K}]W_i^K, [\overleftarrow{V};\overrightarrow{V}]W_i^V)
\end{aligned}
\end{equation}
where $W_i^Q$, $W_i^K$, $W_i^V$ and $W^O$ are parameter projection matrices, which are the same as standard multi-head attention introduced in Equation~\ref{MHAtt}.

\subsection{Integrating Synchronous Bidirectional Attention into NMT} \label{Integrating-sec}

\begin{figure}
	\centering
	\includegraphics[height=10cm]{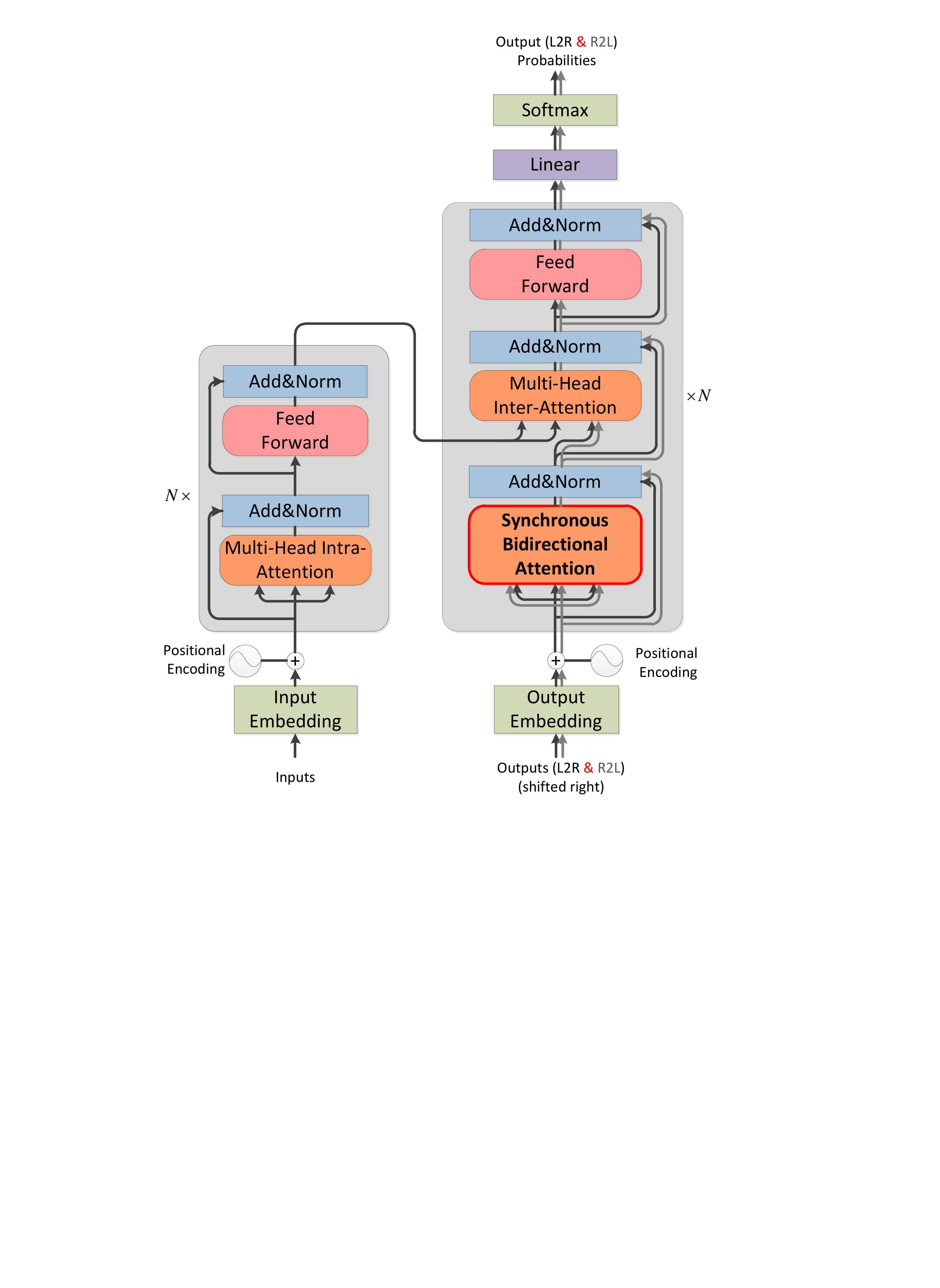}
	\caption{The new Transformer architecture with the proposed synchronous bidirectional multi-head attention network, namely SBAtt.
		The input of decoder is concatenation of forward (L2R) sequence and backward (R2L) sequence. Note that all bidirectional information flow in decoder runs in parallel and only interacts in synchronous bidirectional attention layer.
	}\label{Bi-transformer}
\end{figure}

We apply our synchronous bidirectional attention to replace the multi-head intra-attention in the decoder, as illustrated in Figure~\ref{Bi-transformer}.
The neural encoder of our model is identical to that of the standard Transformer model. From the source tokens, learned embeddings are generated which are then modified by an additive positional encoding. 
The encoded word embeddings are then used as input to the encoder which consists of N blocks each containing two layers: (1) a multi-head attention layer (MHAtt), and (2) a position-wise feed-forward layer (FFN).

The bidirectional decoder of our model is extended from the standard Transformer decoder.
For each layer in the bidirectional decoder, the lowest sub-layer is our proposed synchronous bidirectional attention network, and it also uses residual connections around each of the sublayers, followed by layer normalization.
\begin{equation}
\begin{aligned}
  s^l_d &= {\mbox{LayerNorm}}(s^{l-1} + {\mbox{SBAtt}}(s^{l-1},s^{l-1},s^{l-1})) \\
\end{aligned}
\end{equation}
where $l$ denotes layer depth, subscript $d$ means the decoder-informed intra-attention representation. SBAtt is our proposed synchronous bidirectional attention, and $s^{l-1}$ is equal to $[\overrightarrow{s}^{l-1};\overleftarrow{s}^{l-1}]$ containing forward and backward hidden states.
In addition, the decoder stacks another two sub-layers to seek translation-relevant source semantics to bridge the gap between the source and target language:
\begin{equation}
\begin{aligned}
  s^l_e &= {\mbox{LayerNorm}}(s^{l}_d + {\mbox{MHAtt}}(s^{l}_d, h^N,h^N)) \\
  s^l &= {\mbox{LayerNorm}}(s^l_e + {\mbox{FFN}}(s^l_e))
\end{aligned}
\end{equation}
where MHAtt denotes the multi-head attention introduced in Equation~\ref{MHAtt}, and we use $e$ to denote the encoder-informed inter-attention representation. $h^N$ is the source top layer hidden state, and FFN means feed-forward networks.

Finally, we use a linear transformation and softmax activation to compute the probability of the next tokens based on $s^N=[\overrightarrow{s}^N;\overleftarrow{s}^N]$, namely the final hidden states of forward and backward decoding.
\begin{equation}
\begin{aligned}
    p(\overrightarrow{y}_{j}|\overrightarrow{y}_{<j},\overleftarrow{y}_{<j},x,\theta) = \mbox{Softmax}(\overrightarrow{s}^NW)  \\
    p(\overleftarrow{y}_{j}|\overleftarrow{y}_{<j},\overrightarrow{y}_{<j},x,\theta) = \mbox{Softmax}(\overleftarrow{s}^NW)
\end{aligned}
\end{equation}
where $\theta$ is shared weight for L2R and R2L decoding and $W$ is the weight matrix.

\subsection{Training} \label{Training-sec}
We design a simple yet effective strategy to enable synchronous bidirectional translation within a decoder. 
We separately add the special labels ($\langle l2r \rangle$ and $\langle r2l \rangle$) at the beginning of target sentence ($\overrightarrow{y}$ and $\overleftarrow{y}$) to guide translating from left to right or right to left.
Given a set of training examples $\{x^{(z)}, y^{(z)}\}^Z_{z=1}$, the training algorithm aims to find the model parameters that maximize the likelihood of the training data:
\begin{equation}
\begin{aligned}
J(\theta) = \frac{1}{Z} \sum_{z=1}^Z \sum_{j=1}^M \{ log \ p(\overrightarrow{y}^{(z)}_{j}|\overrightarrow{y}^{(z)}_{<j},\overleftarrow{y}^{(z)}_{<j},x^{(z)},\\ \theta) 
+ \ log \ p(\overleftarrow{y}^{(z)}_{j}|\overleftarrow{y}^{(z)}_{<j}, \overrightarrow{y}^{(z)}_{<j}, x^{(z)},\theta)\}
\end{aligned}
\end{equation}

Similar to asynchronous bidirectional decoding~\cite{zhang2018asynchronous} and bidirectional language models in BERT~\cite{devlin2018bert}, the proposed SB-NMT model also faces the same training problem that the bidirectional decoding would allow the words (the second half of the decoding sequence) to indirectly "see themselves" from the other decoding direction.
To ensure consistency between model training and testing, we construct pseudo references $\overleftarrow{y}_{p}$ ($\overrightarrow{y}_{p}$) for gold $\overrightarrow{y}_{g}$ ($\overleftarrow{y}_{g}$).
More specifically, we first train a L2R model using ($x$, $\overrightarrow{y}_{g}$) and a R2L model using ($x$, $\overleftarrow{y}_{g}$). Then we use the two models to translate source sentences $x$ into pseudo target sentences $\overrightarrow{y}_{p}$ and $\overleftarrow{y}_{p}$ respectively.
Finally, we get two triples $(x, \overrightarrow{y}_{p}, \overleftarrow{y}_{g})$  and $(x, \overrightarrow{y}_{g}, \overleftarrow{y}_{p})$ as our training data.

Once the proposed model is trained, we employ the bidirectional beam search algorithm to predict the target sequence, as illustrated in Figure~\ref{decoding}.
Compared to previous work that usually adopt a two-phase scheme to translate input sentences~\cite{liu2016agreementa,sennrich2017university,zhang2018asynchronous}, our decoding approach is more compact and effective.

\section{Experiments}

We evaluate the proposed model on three translation datasets with different size, including NIST Chinese-English, WMT14 English-German and WMT18 Russian-English translations.

\subsection{Dataset}
For Chinese-English, our training data includes about 2.0 million sentence pairs extracted from the LDC corpus.{\footnote[4]{The corpora includes LDC2000T50, LDC2002T01, LDC2002E18, LDC2003E07, LDC2003E14, LDC2003T17 and LDC2004T07. Following previous work, we also using case-insensitive tokenized BLEU to evaluate Chinese-English which have been segmented by Stanford word segmentation and Moses Tokenizer respectively.}}
 We use NIST 2002 (MT02) Chinese-English dataset as the validation set, NIST 2003-2006 (MT03-06) as our test sets. We use BPE~\cite{Sennrich:2016A} to encode Chinese and English respectively. We learn 30K merge operations and limit the source and target vocabularies to the most frequent 30K tokens.

For English-German translation, the training set consists of about 4.5 million bilingual sentence pairs from WMT 2014.{\footnote[5]{http://www.statmt.org/wmt14/translation-task.html. All preprocessed dataset and vocab can be directly download in tensor2tensor website \url{https://drive.google.com/open?id=0B_bZck-ksdkpM25jRUN2X2UxMm8}.}}
We use newstest2013 as the validation set and newstest2014 as the test set.
Sentences are encoded using BPE, which has a shared vocabulary of about 37000 tokens.
To evaluate the models, we compute the BLEU metric~\cite{P02-1040} on tokenized, true-case output.{\footnote[6]{This procedure is used in the literature to which we compare~\cite{Wu:2016, gehring2017convolutional, vaswani2017attention}.}}

For Russian-English translation, we use the following resources from the WMT parallel data{\footnote[7]{http://www.statmt.org/wmt18/translation-task.html.}}: ParaCrawl corpus, Common Crawl corpus, News Commentary v13 and Yandex Corpus. We do not use Wiki Headlines and UN Parallel Corpus V1.0. The training corpus consists of 14M sentence pairs. We emply the Moses Tokenizer{\footnote[8]{https://github.com/moses-smt/mosesdecoder/blob/mast-er/scripts/tokenizer/tokenizer.perl.}} for precocessing. For subword segmentation, we use 50000 joint BPE operations and choose the most frequent 52000 tokens as vocabularies. We use newstest2017 as the development set and the newtest2018 as the test set.

\begin{table}
	\centering
	\begin{tabular}{|l|l||c|c|c|}
		\hline
		\multicolumn{2}{|c||}{Fusion}       &    $\lambda$ =0.1  &$\lambda$ =0.5   &$\lambda$ =1.0   \\
		\hline
		\hline
		\multicolumn{2}{|c||}{Linear}                  &     51.05     & 50.71 &  46.98    \\
		\hline
		\multirow{2}{*}{Nonlinear}        &\textit{tanh} &     50.99   &     50.72    &   50.96    \\  \cline{2-5}
		&\textit{relu} &     50.79   &     50.57      &   50.71         \\
		\hline
		\multicolumn{2}{|c||}{Gate}                    &      \multicolumn{3}{c|}{50.51}        \\
		\hline
	\end{tabular}
	\caption{Experiment results on the development set using different fusion mechanism with different $\lambda$s.} \label{acg-table}
\end{table}

\begin{table*}
	\centering
	\begin{tabular}{l|c|cccc|cc}
		\hline
		Model                 &  DEV      &MT03   &      MT04   &      M05     &      MT06       & AVE      &  $\Delta$ \\
		\hline
		\hline
		Moses             &  37.85     & 37.47     &    41.20     &       36.41    &    36.03        &     37.78   & -9.41   \\
		RNMT               &  42.43   &  42.43    &    44.56      &     41.94    &   40.95       &     42.47    &   -4.72\\
		Transformer     &  48.12   &  47.63    &    48.32     &       47.51     &   45.31       &     47.19    & -\\
		Transformer (R2L)  &   47.81    &  46.79     &    47.01      &      46.50   &   44.13       &     46.11    &  -1.08\\
		Rerank-NMT          &  49.18   &  48.23       & 48.91    &     48.73    & 46.51       &     48.10 &  +0.91 \\
		ABD-NMT           &   48.28 &  49.47     &       48.01        &       48.19        &      47.09    &   48.19    & +1.00 \\
		\hline
		\hline
		Our Model         &  \textbf{50.99}    &\textbf{51.87}        &     \textbf{51.50}   &        \textbf{51.23}    &    \textbf{49.83}    &  \textbf{51.11}  & \textbf{+3.92}   \\		
		\hline
	\end{tabular}
	\caption{Evaluation of translation quality for Chinese-English translation tasks using case-insensitive BLEU scores.
		All results of our model are significantly better than Transformer and Transformer (R2L) (p $<$ 0.01).
	} \label{CH-EN}
\end{table*}

\subsection{Setting}
We build the described models by modifying the tensor2tensor{\footnote[9]{https://github.com/tensorflow/tensor2tensor.}} toolkit for training and evaluating.
For our bidirectional Transformer model, we employ the Adam optimizer with $\beta_1$=0.9, $\beta_2$=0.998, and $\epsilon$=$10^{-9}$. We use the same warmup and decay strategy for learning rate as \newcite{vaswani2017attention}, with 16,000 warmup steps.
During training, we employ label smoothing of value $\epsilon_{ls}$=0.1.
For evaluation, we use beam search with a beam size of $k$=4 (For SB-NMT, we use two L2R and R2L hypotheses respectively.) and length penalty $\alpha$=0.6.
Additionally, we use 6 encoder and decoder layers, hidden size $d_{model}$=1024, 16 attention-heads, 4096 feed forward inner-layer dimensions, and $P_{dropout}$=0.1.
Our settings are close to \emph{transformer\_big} setting as defined in \newcite{vaswani2017attention}.
We employ three Titan Xp GPUs to train English-German and Russian-English translation, and one GPU for Chinese-English translation pairs. In addition, we use a single model obtained by averaging the last 20 checkpoints for English-German and Russian-English and do not perform checkpoint averaging for Chinese-English.

\subsection{Baselines}

We compare the proposed model against the following state-of-the-art SMT and NMT systems{\footnote[10]{For fair comparison, Rerank-NMT and ABD-NMT are based on strong Transformer models.}}:

\begin{itemize}
	\item \textbf{Moses}: an open source phrase-based SMT system with default configuration and a 4-gram language model trained on the target portion of training data.
	\item \textbf{RNMT}~\cite{Luong:2015A}: it is a  state-of-the-art RNN-based NMT system with default setting.
	\item \textbf{Transformer}: it has obtained the state-of-the-art performance on machine translation, which predicts target sentence from left to right relying on self-attention~\cite{vaswani2017attention}.
	\item \textbf{Transformer (R2L)}: it is a variant of Transformer that generates translation in a right-to-left direction.
	\item \textbf{Rerank-NMT}: Via exploring the agreement on left-to-right and right-to-left NMT models, \cite{liu2016agreementa,W16-2323} first run beam search for forward and reverse models independently to obtain two k-best lists, and then re-score the union of two k-best lists (k=10 in our experiments) using the joint model (adding logprobs) to find the best candidate.
	\item \textbf{ABD-NMT}: it is an asynchronous bidirectional decoding for NMT, which equipped the conventional attentional encoder-decoder NMT model with a backward decoder~\cite{zhang2018asynchronous}.
	ABD-NMT adopts a two-phrase decoding scheme: (1) use backward decoder to generate reverse sequence states; (2) perform beam search on the forward decoder to find the best translation based on encoder hidden states and backward sequence states.
\end{itemize}

\subsection{Results on Chinese-English Translation}

\textbf{Effect of Fusion Mechanism}
We first investigate the impact of different fusion mechanisms with different $\lambda$s on the development set.
As shown in Table~\ref{acg-table}, we find that linear interpolation is sensitive to parameters $\lambda$.
Nonlinear interpolation, which is more robust than linear interpolation, achieves the best performance when we use $tanh$ with $\lambda$=0.1. Compared to gate mechanism, nonlinear interpolation is much simpler and needs less parameters. Therefore, we will use nonlinear interpolation with  $tanh$ and $\lambda$=0.1 for all experiments thereafter.

\textbf{Translation Quality}
Table~\ref{CH-EN} shows translation performance for Chinese-English.
Specifically, the proposed model significantly outperforms Moses, RNMT, Transformer, Transformer (R2L), Rerank-NMT and ABD-NMT by 13.23, 8.54, 3.92, 4.90, 2.91, 2.82 BLEU points, respectively.
Compared to Transformer and Transformer (R2L), our model exhibits much better performance. These results confirm our hypothesis that the two directions are mutually beneficial in bidirectional decoding. 
Furthermore, compared to Rerank-NMT in which two decoders are relatively independent and ABD-NMT where only the forward decoder can rely on a backward decoder, our proposed model achieves substantial improvements over them on all test sets, which indicates that joint modeling and optimizing with left-to-right and right-to-left decoding behaves better in leveraging bidirectional decoding.

\subsection{Results on English-German Translation}

We further demonstrate the effectiveness of our model in WMT14 English-German translation tasks, and we also display the performances of some competitive models including GNMT~\cite{Wu:2016}, Conv~\cite{gehring2017convolutional}, and AttIsAll~\cite{vaswani2017attention}.
As shown in Table~\ref{EN-GE}, our model also significantly outperforms others and gets an improvement of 1.49 BLEU points than a strong Transformer model. 
Moreover, our SB-NMT model establishes a state-of-the-art BLEU score of 29.21 on the WMT14 English-German translation task.

\begin{table}
	\centering
	\begin{tabular}{l|c}
		\hline
		Model              &    TEST               \\
		\hline
		\hline
		GNMT$\ddagger$~\cite{Wu:2016}                 &         24.61            \\
		Conv$\ddagger$~\cite{gehring2017convolutional}     &           25.16          \\
		AttIsAll$\ddagger$~\cite{vaswani2017attention}     &      28.40        \\
		\hline
		\hline
		Transformer\footnotemark[11]   &       27.72     \\
		Transformer (R2L)  &      27.13          \\
		Rerank-NMT          &     27.81      \\
		ABD-NMT           &     28.22      \\ 
		\hline
		\hline
		Our Model         &     \textbf{29.21}     \\		
		\hline
	\end{tabular}
	\caption{Results of WMT14 English-German translation using case-sensitive BLEU. 
		Results with $\ddagger$ mark are taken from the corresponding papers.
	} \label{EN-GE}
\end{table}
\footnotetext[11]{The BLEU scores for Transformer model are our reproduced results. Similar to footnote 7 in \cite{P18-1008}, our performance is slightly lower than those reported in ~\cite{vaswani2017attention}. Additionally, we only use 3 GPUs for English-German, whereas most papers employ 8 GPUs for model training.}

\begin{table}
	\centering
	\begin{tabular}{l|cc}
		\hline
		Model              &  DEV &  TEST               \\
		\hline
		\hline
		Transformer       &  35.28     &   31.02     \\
		Transformer (R2L)  &  35.22     &    30.57          \\
		Our Model         &   \textbf{36.38}    &   \textbf{32.06}     \\	
		\hline
	\end{tabular}
	\caption{Results of WMT18 Russian-English translation using case-insensitive tokenized BLEU.
	} \label{Ru-En}
\end{table}

\subsection{Results on Russian-English Translation}

Table~\ref{Ru-En} shows the results of large-scale WMT18 Russian-English translation, and our approach still significantly outperforms the state-of-the-art Transformer model in development and test sets by 1.10 and 1.04 BLEU points respectively.
Note that the BLEU score gains of English-German and Russian-English are not as significant as that on Chinese-English. The underlying reasons, which have also been mentioned in \newcite{Shen:2016} and \newcite{zhang2018asynchronous}, are that
(1) the Chinese-English datasets contain four reference translations for each source sentence while the English-German and Russian-English datasets only have single reference;
(2) English is more distantly related to Chinese than German and Russian, leading to the predominant improvements for Chinese-English translation when leveraging bidirectional decoding.

\subsection{Analysis}

\begin{table}
	\centering
	\begin{tabular}{l|c|cc}
		\hline
		\multirow{2}{*}{Model} &  \multirow{2}{*}{Param}   &   \multicolumn{2}{c}{Speed} \\
		&                               &    \emph{Train}   & \emph{Test}     \\
		\hline
		\hline
		Transformer       &  207.8M &    2.07       &  19.97 \\
		Transformer (R2L)  &  207.8M &    2.07          &  19.81  \\
		Rerank-NMT         &  415.6M&    1.03       & 6.51\\
		ABD-NMT            &  333.8M &    1.18          & 7.20 \\
		\hline
		\hline
		Our Model          &  207.8M &      1.26   &    17.87 \\
		\hline
	\end{tabular}
	\caption{Statistics of parameters, training and testing speeds. \emph{Train} denotes the number of global training steps processed per second at the same batch-size sentences; \emph{Test} indicates the amount of translated sentences in one second.} \label{params}
\end{table}

We conduct analyses on Chinese-English translation, to better understand our model from different perspectives.

\textbf{Parameters and Speeds}
In contrast to the standard Transformer, our model does not increase any parameters except for a hyper-parameter $\lambda$, as shown in Table~\ref{params}.
Rerank-NMT needs to train two sets of NMT models, so its parameters are doubled.
The parameters of ABD-NMT are 333.8M since it has two decoders containing a backward decoder and a forward decoder.
Hence, our model is more compact because it only has a single encoder-decoder NMT model.

We also show the training and testing speed of our model and baselines in Table~\ref{params}. 
During training, our model performs approximately 1.26 training steps per second, which is faster than Rerank-NMT and ABD-NMT.
When it comes to decoding procedure, the decoding speed of our model is 17.87 sentences per second with batch size 50, which is two or three times faster than Rerank-NMT and ABD-NMT.

\begin{figure}

	\setlength{\belowcaptionskip}{-0.2cm}
    \centering
    \includegraphics[width=7cm]{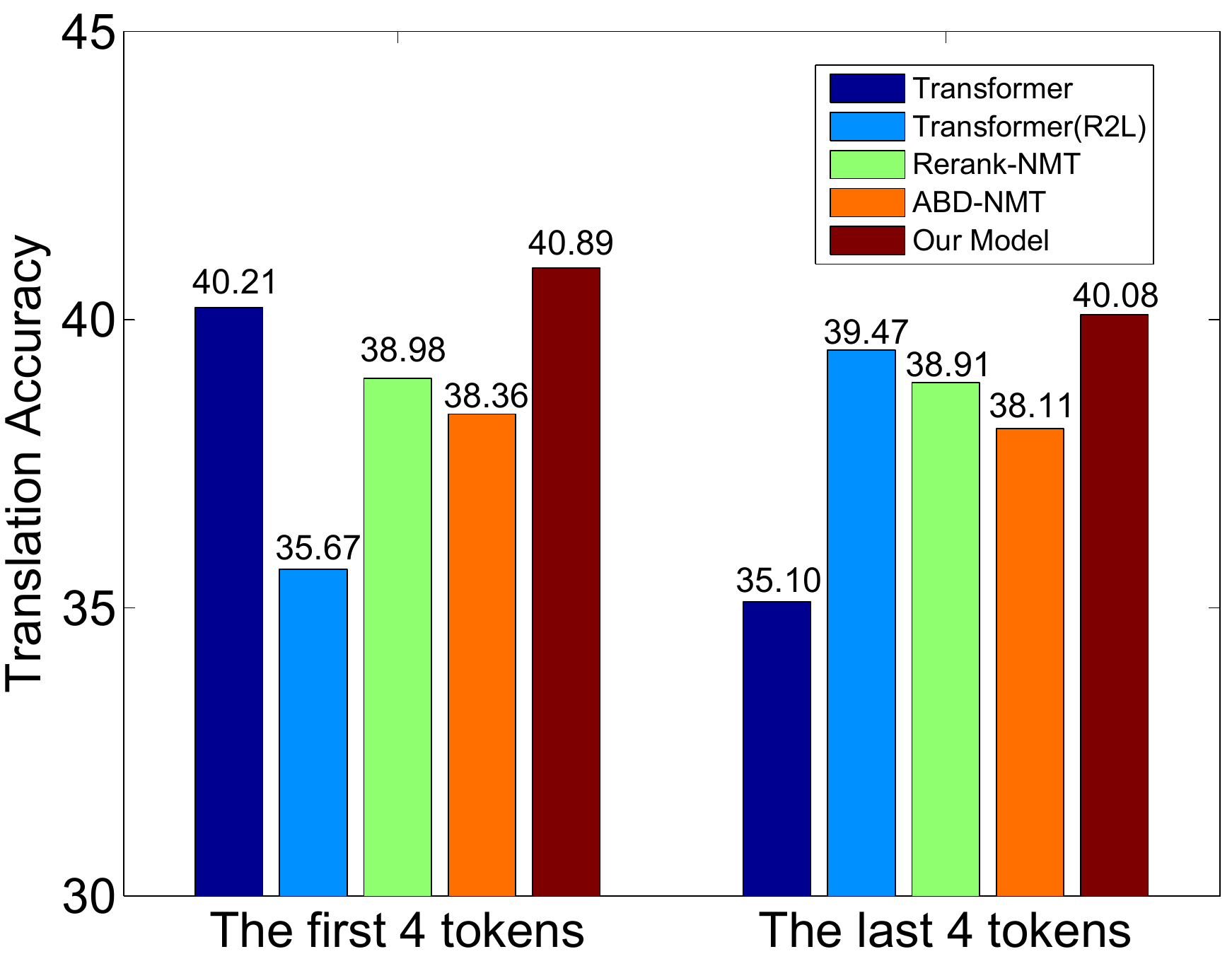}
    \caption{Translation accuracy of the first and last 4 tokens for Transformer, Transformer (R2L), Rerank-NMT, ABD-NMT and our proposed model.
    }\label{tacl-acc-3}
\end{figure}

\textbf{Effect of Unbalanced Outputs}
According to Table~\ref{trans-acc}, L2R usually does well on predicting the left-side tokens of target sequences, while R2L usually performs well on the right-side tokens.
Our central idea is combine the advantage of left-to-right and right-to-left modes. To test our hypothesis, we further analyze the translation accuracy of Rerank-NMT, ABD-NMT, and our model, as shown in Figure~\ref{tacl-acc-3}. 
Rerank-NMT and ABD-NMT can alleviate the unbalanced output problem, but fail to improve prefix and suffix accuracies at the same time. 
The experimental results demonstrate that our model can balance the outputs, and gets the best translation accuracy for both the first 4 words and the last 4 words.
Note that our model chooses from L2R output or R2L output as final results according to their model probabilities, and the left-to-right decoding contributes 58.6\% on test set.

\begin{figure}
	\setlength{\abovecaptionskip}{0cm}
	\setlength{\belowcaptionskip}{-0.2cm}
	\centering
	\includegraphics[height=6cm]{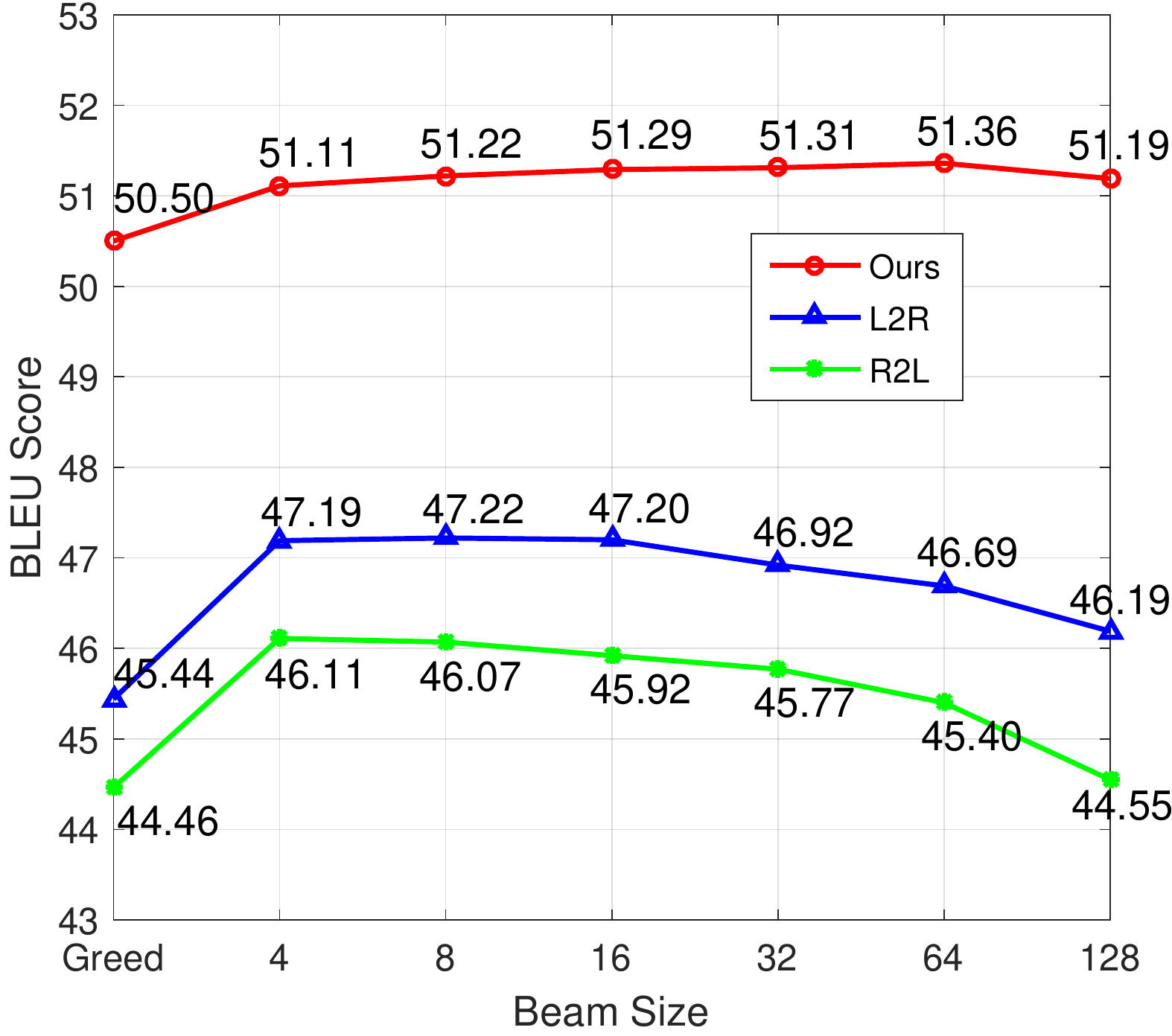}
	\caption{ Translation qualities (BLEU score) of our L2R, R2L and our SB-NMT model as beam size becomes larger\footnotemark[12]. 
	}\label{tacl_beam}
\end{figure}
\footnotetext[12]{For greedy search in SB-NMT, it has one item L2R decoding and one item R2L decoding. In other words, its beam size is equal to 2 compared to conventional beam search decoding.}

\begin{figure*}
	\setlength{\abovecaptionskip}{0cm}
	\setlength{\belowcaptionskip}{-0.1cm}
	\centering
	\includegraphics[height=6.5cm]{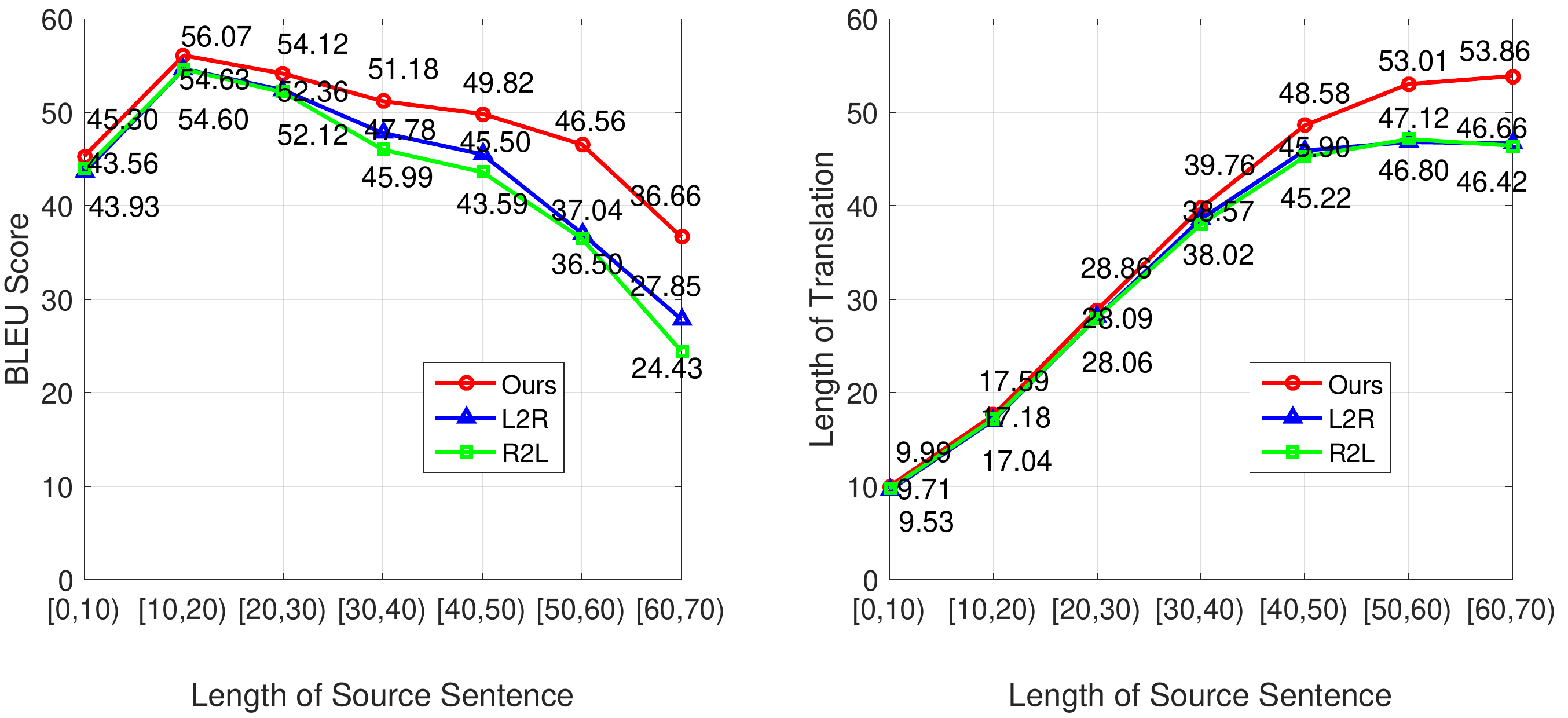}
	\caption{Performance of translations on the test set with respect to the lengths of the source sentences.
	}\label{tacl_len_2}
\end{figure*}

\textbf{Effect of Varying Beam Size}
We observe that beam search decoding only improves translation quality for narrow beams and degrades translation quality when exposed to a larger search space for L2R and R2L decoding as illustrated in Figure~\ref{tacl_beam}. 
Additionally, the gap between greedy search and beam search is significant and can be up to about 1-2 BLEU points.
\newcite{koehn2017six} also demonstrate these phenomena in eight translation directions.

As for our SB-NMT model, we investigate the effect of different beam sizes k, as shown by the red line of Figure~\ref{tacl_beam}. 
Compared to conventional beam search, where worse translations are found beyond an optimal beam size setting (e.g., in the range of 4-32), the translation quality of our proposed model remains stable as beam size becomes larger.
We attribute this to the ability of the combined objective to model both history and future translation information.

\textbf{Effect of Long Sentences}
A well-known flaw of NMT models is the inability to properly translate long sentences.
We follow \newcite{Bahdanau:2015} to group sentences of similar lengths together and compute a BLEU score per group (left picture). Figure~\ref{tacl_len_2} shows the BLEU score and the averaged length of translations for each group (right picture).
Transformer and Transformer (R2L) perform very well on short source sentences, but degrade on long source sentences. 
Our model can alleviate this problem by taking advantage of both history and future information.
In fact, incorporating synchronous bidirectional attention boosts translation performance on all source sentence groups.

\textbf{Comparison to Data-Enhanced NMT}
In the training setup, we have obtained pseudo L2R and R2L references ($\overrightarrow{y}_{p}$ and $\overleftarrow{y}_{p}$) by using L2R and R2L models respectively.
Here, we first compare our proposed model with NMT enhanced by pseudo data, and further explore the data utilization of SB-NMT by using combined data strategy (six triples data, that is, ($\overrightarrow{y}_{g}$, $\overleftarrow{y}_{p}$), (reversed $\overleftarrow{y}_{p}$, $\overleftarrow{y}_{g}$),
($\overrightarrow{y}_{p}$, $\overleftarrow{y}_{g}$), ($\overrightarrow{y}_{g}$, reversed $\overrightarrow{y}_{p}$),
($\overrightarrow{y}_{p}$, $\overleftarrow{y}_{p}$), and (reversed $\overleftarrow{y}_{p}$, reversed $\overrightarrow{y}_{p}$)).
As shown in Table~\ref{pesudo-data}, we find that data-enhanced Transformer outperforms the original Transformer, but still behaves worse than our proposed model.
Furthermore, by making full use of training data, our model (six triple data) significantly improves the translation quality by 1.03 BLEU points than the original set (two triples data).

\begin{table}
	\setlength{\belowcaptionskip}{-0.2cm}
	\centering
	\begin{tabular}{l|c}
		\hline
		Model                 &  TEST   \\
		\hline
		\hline
		Transformer (standard $\overrightarrow{y}_{g}$)             &  47.17  \\
		SB-NMT (two triples data)               &  51.11  \\
		\hline
		Transformer ($\overrightarrow{y}_{g}$ + $\overrightarrow{y}_{p}$)     &  49.48  \\
		Transformer ($\overrightarrow{y}_{g}$ +  $\overrightarrow{y}_{p}$ + $\overleftarrow{y}_{p}$ )     &  49.99  \\
		\hline
		SB-NMT (six triples data)         &  \textbf{52.14}   \\		
		\hline
	\end{tabular}
	\caption{Chinese-English BLEU scores of standard Transformer enhanced with pseudo data, and our SB-NMT model with combined data strategy.
	} 
	\label{pesudo-data}
\end{table}

\textbf{Subjective Evaluation}
We follow \newcite{Tu:2016} to conduct a subjective evaluation to validate the benefit of the synchronous bidirectional decoder, as shown in Table~\ref{sub-evaluation}.
Four human evaluators are asked to evaluate the translations of 100 source sentences, which are randomly sampled from the test sets without knowing which system the translation is selected from.
These 100 source sentences have 2712 words. We evaluate over- or under-translation based on the number of source words which are dropped or repeated in translation{\footnote[13]{For our SB-NMT model, 2 source words are over-translated and 147 source words are under-translated. Additionally, it is interesting to combine with better scoring methods and stopping criteria~\cite{D18-1342} to strengthen the baseline and our model in the future.}}, though we use subword~\cite{Sennrich:2016A} in training and inference.
Transformer and Transformer (R2L) suffer from serious under-translation problems with 7.85\% and 7.81\% errors.
Our proposed model alleviates the under-translation problems by exploiting the combination of left-to-right and right-to-left decoding directions, reducing 30.6\% of under-translation errors.
It should be emphasized that the proposed model is especially effective for alleviating under-translation problem, which is a more serious translation problem for Transformer systems as seen in Table~\ref{sub-evaluation}.

\begin{table}
	\setlength{\belowcaptionskip}{-0.2cm}
	\centering
	\begin{tabular}{l|c|c|c|c}
		\hline
		\multirow{2}{*}{Model}    &   \multicolumn{2}{c|}{Over-Trans}  &   \multicolumn{2}{c}{Under-Trans} \\ \cline{2-5}
		&    Ratio   & $\Delta$     &    Ratio   & $\Delta$   \\
		\hline
		L2R   &     0.07\%       &   -   & 7.85\%      &  -   \\
		R2L   &      0.14\%       &  -   & 7.81\%       & -   \\
		Ours  &   0.07\%     &  -0.00\%    &  5.42\%    &   -30.6\%  \\
		\hline
	\end{tabular}
	\caption{Subjective evaluation on over-translation and under-translation for Chinese-English.
		Ratio denotes the percentage of source words which are over- or under-translated, $\Delta$ indicates relative improvement.} \label{sub-evaluation}
\end{table}

\begin{table*}[!ht]
	\setlength{\belowcaptionskip}{-0.3cm}
	\centering
	\begin{tabular}{|p{1.5cm}|p{13cm}|}
		\hline
		Source &  \textcolor{red}{ \dashuline{捷克 \ 总统 \ 哈维 \ 卸任}} \ \textcolor{blue}{\uwave{新 \ 总统 \ 仍 \ 未 \ 确定}} \\
		\hline
		Reference & czech president havel steps down while new president still not chosen  \\
		\hline
		L2R & \textcolor{red}{\dashuline{czech president leaves office}}   \\
		\hline
		R2L & \textcolor{blue}{\uwave{the outgoing president of the czech republic is still uncertain}} \\
		\hline
		Ours &  \textcolor{red}{\dashuline{czech president havel leaves office}} , \textcolor{blue}{\uwave{new president yet to be determined}}   \\
		\hline
		\hline
		Source &  \textcolor{red}{ \dashuline{他们 \ 正在 \ 研制 \ 一 \ 种 \ 超大型}} \ 的 \ \textcolor{blue}{\uwave{叫做 \ 炸弹 \ 之 \ 母 \ 。}} \\
		\hline
		Reference & they are developing a kind of superhuge bomb called the mother of bombs .  \\
		\hline
		L2R & \textcolor{red}{ \dashuline{they are developing a super , big}} , mother , called the bomb .   \\
		\hline
		R2L &  they are working on a much larger mother \textcolor{blue}{\uwave{called the mother of a bomb .}} \\
		\hline
		Ours &  \textcolor{red}{ \dashuline{they are developing a super-large scale}} , \textcolor{blue}{\uwave{called the mother of the bomb .}}  \\
		\hline
	\end{tabular}
	\caption{Chinese-English translation examples of Transformer decoding in left-to-right and right-to-left way, and our proposed models.
		L2R performs well in \textcolor{red}{\protect\dashuline{the first half sentence}}, whereas R2L translates well in \textcolor{blue}{\uwave{the second half sentence}.} } \label{example}
\end{table*}

\textbf{Case Study}
Table~\ref{example} gives three examples to show the translations of different models, in order to better understand how our model outperforms others.
We find that Transformer produces translations with good prefixes (\textcolor{red}{red line} or \dashuline{dotted line}), while Transformer (R2L) generates translations with better suffixes (\textcolor{blue}{blue line} or \uwave{wave line}).
Therefore, they are often unable to translate the whole sentence precisely.
In contrast, the proposed approach can make full use of bidirectional decoding and remedy the errors in these cases.

\section{Related Work}

Our research is built upon a sequence-to-sequence model~\cite{vaswani2017attention}, but it is also related to future modeling and bidirectional decoding. We discuss these topics in the following.

\textbf{Future Modeling} Standard neural sequence decoders generate target sentences from left to right,
and it has been proven to be important to establish the direct information flow between current predicting word and previous generated words~\cite{zhou2017,vaswani2017attention}.
However, current methods still fail to estimate some desired information in the future.
To address this problem, reinforcement learning methods have been applied to predict future properties~\cite{li2017learning,bahdanau2017actor,he2017decoding}.
\newcite{N18-1125} presented a target foresight based attention which uses the POS tag as the partial information of a target foresight word to improve alignment and translation.
Inspired by the human cognitive behaviors, \newcite{NIPS2017_6775} proposed a deliberation network, which leverages the global information by observing both back and forward information in sequence decoding through a deliberation process.
\newcite{Q18-1011} introduced two additional recurrent layers to model translated past contents and untranslated future contents.
The most relevant models in future modeling are twin networks~\cite{serdyuk2018twin}, which encourage the hidden state of the forward network to be close to that of the backward network used to predict the same token.
However, they still used two decoders and the backward network contributes nothing during inference.
Along the direction of future modeling, we introduce a single synchronous bidirectional decoder, where forward decoding can be used as future information for backward decoding, and vice versa.

\textbf{Bidirectional Decoding} In SMT, many approaches explored backward language models or target-bidirectional decoding to capture right-to-left target-side contexts for translation~\cite{C02-1050,D09-1117,N13-1002}. To address the issue of unbalanced outputs, \newcite{liu2016agreementa} proposed an agreement model to encourage the agreement between L2R and R2L NMT models. Similarly, some work attempted to re-rank the left-to-right decoding results by right-to-left decoding, leading to diversified translation results~\cite{W16-2323,D17-1014,tan2017xmu,sennrich2017university,liu2018,W18-6408}.
Recently, \newcite{zhang2018asynchronous} proposed asynchronous bidirectional decoding for NMT, which extended the conventional attentional encoder-decoder framework by introducing a backward decoder. Additionally, both \newcite{Niehues:2016} and \newcite{P17-2060} combined the strengths of NMT and SMT, which can also be used to combine the advantages of bidirectional translation texts~\cite{zhang2018asynchronous}.
Compared to previous methods, our method has the following advantages:
(1) We use a single model to achieve the goal of synchronous left-to-right and right-to-left decoding.
(2) Our model can leverage and combine the two decoding directions in every layer of the Transformer decoder, which can run in parallel.
(3) By using synchronous bidirectional attention, our model is an end-to-end joint framework and can optimize L2R and R2L decoding simultaneously.
(4) Compared to two-phase decoding schemes in previous work, our decoder is more compact and faster.

\section{Conclusions and Future Work}

In this paper, we propose a synchronous bidirectional NMT model that performs bidirectional decoding simultaneously and interactively.
The bidirectional decoder, which can take full advantage of both history and future information provided by bidirectional decoding states, predicts its outputs using left-to-right and right-to-left directions at the same time.
To the best of our knowledge, this is the first attempt to integrate synchronous bidirectional attention into a single NMT model.
Extensive experiments demonstrate the effectiveness of our proposed model.
Particularly, our model respectively establishes state-of-the-art BLEU scores of 51.11 and 29.21 on NIST Chinese-English and WMT14 English-German translation tasks.
In future work, we plan to apply this framework to other tasks, such as sequence labeling, abstractive summarization and image captioning.
Additionally, it is interesting to reduce the training cost by adding noise in the target sentence and using fine-tune technology.

\section*{Acknowledgments}

We would like to thank the anonymous reviewers as well as the Action Editor, George Foster, for insightful comments and suggestions. 
The research work has been funded by the Natural Science Foundation of China under Grant No. 61673380.
This work is also supported by grants from NVIDIA NVAIL program.

\bibliography{tacl2018}
\bibliographystyle{acl_natbib}

\end{CJK*}
\end{document}